\pdfoutput=1 
\documentclass[11pt]{article}
\usepackage[hyperref, final]{eacl2023}
\usepackage{nert}
\usepackage{times}
\usepackage{xcolor}
\usepackage{latexsym}
\usepackage{empheq}
\usepackage{bbm}
\usepackage{comment}

\newcommand{\lastrow}[0]{$\hspace{0.5em}\hookrightarrow$}

\newcommand{\samel}[0]{\multicolumn{1}{l}{\space\hspace{0.5em}\vline}}

\newcommand{\lr}[2]{#1e--#2}

\definecolor{Purple4}{HTML}{550055}

\usepackage{microtype}

\title{PECO: Examining Single Sentence Label Leakage in Natural Language Inference Datasets through Progressive Evaluation of Cluster Outliers
}

\author{Michael Saxon, Xinyi Wang, Wenda Xu, William Yang Wang \\
  University of California, Santa Barbara\\
  Department of Computer Science \\
  \eml{saxon@ucsb.edu}, \eml{xinyi\_wang@ucsb.edu}, \eml{wendaxu@ucsb.edu}, \eml{william@cs.ucsb.edu}
}

\date{}

\begin{document}

\maketitle
\begin{abstract}
Building natural language inference (NLI) benchmarks that are both challenging for modern techniques, and free from shortcut biases is difficult. Chief among these biases is \textit{single sentence label leakage}, where annotator-introduced spurious correlations yield datasets where the logical relation between (premise, hypothesis) pairs can be accurately predicted from only a single sentence, something that should in principle be impossible. We demonstrate that despite efforts to reduce this leakage, it persists in modern datasets that have been introduced since its 2018 discovery. To enable future amelioration efforts, introduce a novel model-driven technique, the progressive evaluation of cluster outliers (PECO) which enables both the objective measurement of leakage, and the automated detection of subpopulations in the data which maximally exhibit it.
\end{abstract}

\section{Introduction}

Natural language inference (NLI) is a fundamentally pairwise task, wherein a logical relation between two statements is predicted.
Progress on NLI benchmarks is an important proxy for advancements in natural language reasoning by machines.
Models are trained to process the two statements simultaneously, in a \textit{paired sentence condition} (PSC).
Unfortunately many modern NLI datasets exhibit 
\textbf{single sentence label leakage}.
When this leakage is present, models are able to accurately predict the pairwise relation encoded by the labels in a \textit{single sentence condition} (SSC)---where the model is only shown one of the statements \cite{poliak2018hypothesis}.
This is a serious problem, rendering NLI datasets' capture of reasoning questionable, and limiting the robustness of models trained on them. 

NLI is formalized as predicting a relation $r\in$\{neutral, entail, contradict\} from a pair of sentences (premise $s_1$ and hypothesis $s_2$).
An ideal NLI benchmark without single sentence label leakage will have distribution of $r$ that is conditionally dependent on the \textbf{pair} of sentences, but independent from either individual sentence  \cite{wang2021counterfactual}. In practice this 
is difficult to achieve, particularly when constructing usefully large datasets.

\begin{figure}[t!]
    \centering
    \includegraphics[width=0.95\linewidth]{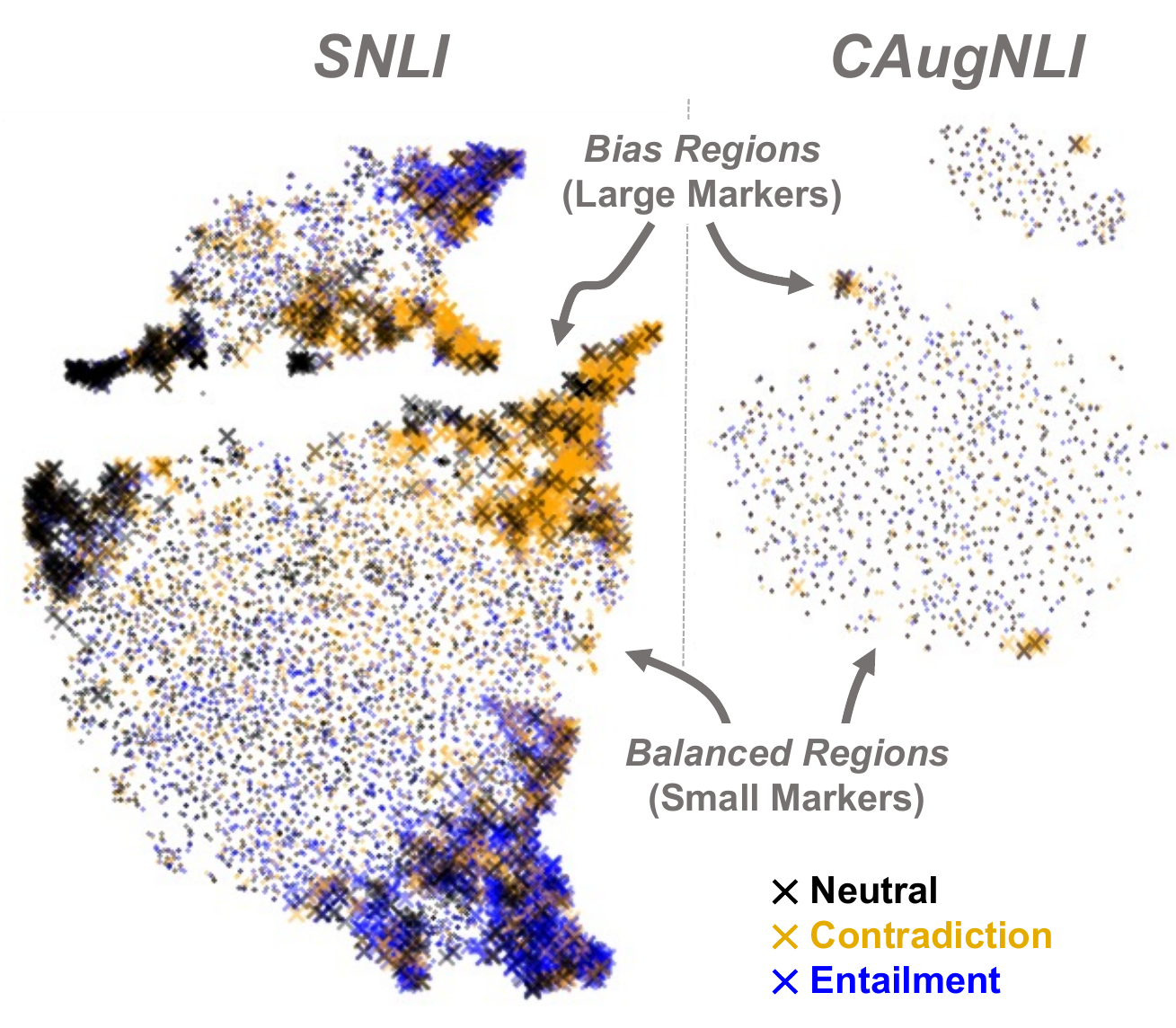}
    \caption{A T-SNE projection of the SNLI and CAugNLI test sets in PECO's model-driven \textit{single sentence condition} (SSC) embedding space, showing \textcolor{blue}{\textbf{entailment}}-, \textcolor{orange}{\textbf{contradiction}}-, and \textbf{neutral}-labeled samples. 
    This model was trained on the paired-sentence condition (PSC) where the relation between sentences in a sample are observable.
    In a leakage-free dataset, a model should be unable to separate subpopulations that disproportionately exhibit one label class in the SSC. 
    Local regions exhibiting an imbalanced label distribution in this subpopulation are considered ``biased regions'' and plotted with large markers.
    SNLI exhibits large, continuous regions in the hypothesis only space disproportionately exhibiting the same label, compared to CAugNLI. Accordingly, SNLI exhibits higher single sentence label leakage than CAugNLI (\autoref{tab:results}).
    }
    \label{fig:clust-snli-test}
    \vspace{-.5cm}
\end{figure}


\begin{table*}[t!]
\small
    \centering
    \begin{tabular}{lllllrrr}
        \toprule
        Dataset & Authors (Year) & Seed sentences from & Language & Imbalance on & \# Train & \# Dev & \# Test \\
        \hline
        SICK & \citet{marelli2014sick} & Image+Video Captions  & En & $s_2$ & 4.4k & 0.5k & 4.9k \\
        \vspace{0.3em}SNLI & \citet{bowman2015snli} & Image Captions + KB & En & $s_2$ & 550k & 10k & 10k \\
        MNLI & \citet{williams2018mnli} & Multiple Genre & En & $s_2$ & 393k & 20k  & 20k\\
        \lastrow MNLI-m & \samel  & \samel  & \samel & \samel & --- & 10k & 10k \\
        \vspace{0.3em}\lastrow MNLI-u & \samel & \samel  & \samel & \samel & --- & 10k & 10k \\
        XNLI & \citet{conneau2018xnli} & MNLI & 14 Langs. & $s_2$ & --- & 70k & 35k \\
        \vspace{0.3em}FEVER$_\text{NLI}$ & \citet{nie2019combining} & Wikipedia & En & $s_1$ & 208k & 20k & 20k \\
        ANLI & \citet{nie2020adversarial} & Wikipedia+HotpotQA & En & $s_2$ & 163k & 3.2k & 3.2k \\
        \lastrow A1 & \samel & \samel & \samel & \samel & 17k & 1k & 1k \\
        \lastrow A2 & \samel & \samel & \samel & \samel & 45k & 1k & 1k \\
        \vspace{0.3em}\lastrow A3 & \samel & \samel & \samel & \samel & 100k & 1.2k & 1.2k \\
        OCNLI & \citet{ocnli} & Multiple Genre & Zh & $s_2$ & 47k & 3k & 3k \\
        CAugNLI & \citet{kaushik2020learning} & IMDb+SNLI & En & $s_2$ \& $s_1$ & 8.3k & 1k & 8.3k \\
        SNLI$_{\text{debiased}}$ & \citet{wu2022generating} & SNLI, generated & En & $s_2$ \& $s_1$ & 1.14M & --- & --- \\
        MNLI$_{\text{debiased}}$ & \samel & MNLI, generated & \samel & \samel & 744k & --- & --- \\
        \bottomrule
    \end{tabular}
    \caption{Information on the NLI datasets we compare in this study. MNLI has ``matched'' (m) and ``unmatched'' (u) test sets that we evaluate separately. The ANLI dataset is decomposed into partitions ``A1,'' ``A2,'' and ``A3''. A Vertical line denotes identical value in cell as above (sub-elements of same dataset).}
    \vspace{-2ex}
    \label{tab:dataset_info}
\end{table*}

Most large-scale NLI datasets are produced by sourcing \textit{seed} sentences from an existing text population to serve as initial premises or hypotheses. Each seed sentence is then 
assigned one or more relations $r$, which annotators 
use to write new \textit{elicited} sentences satisfying each selected relation relative to the seed.
Many datasets exclusively build either the hypothesis or premise population from seed sentences, leaving the other exclusively elicited. 

Systematic, shared biases in the words, sentence structures, or ideas that crowdworkers consider when given a logical relation, coupled with the exclusively elicited nature of one of the sentence populations, then drive relation leakage \cite{gururangan-2018-annotation}. 
For example, a slight preference for 
words like ``not'' or ``doesn't'' when given \textit{contradict} as opposed to \textit{entail} would lead to a bias in the n-gram distribution between the classes in the SSC. Simple heuristics inspired by these findings can produce challenging test sets that hobble models trained on these biased datasets \cite{mccoy2020right}, but they require manual guesswork, don't generalize, and may miss higher-level, more nuanced semantic shortcuts and biases.

These ``leakage features'' encoded in the elicited sentences are visible to NLI models \cite{zhang_selection_2019}, enabling them to ``cheat'' by attending to them as shortcuts rather than logical correspondences between the two sentences, calling into question the appropriateness of NLI datasets as benchmarks for language understanding \cite{bowman2021will}. In this work we rigorously analyze this problem of single sentence relation leakage in both popular and recent NLI datasets using novel techniques to enable targeted interventions and create higher quality future resources.





Further NLI datasets have been proposed to tackle these problems using machine-in-the-loop adversarial sentence elicitation, \cite{nie2020adversarial}, counterfactual augmentation \cite{kaushik2020learning}, 
and learning dynamic-based debiasing \cite{wu2022generating}. 
These datasets are purported to provide more challenging generalization scenarios for NLI models
to better test logical reasoning capabilities. 
One big question remains---\textit{have these techniques actually eliminated relation leakage biases?}



In this work, we demonstrate the following:

\paragraph{New NLI datasets still exhibit single sentence relation leakage.} We compare SSC performance for 10 NLI datasets (including those previously assessed by \citet{poliak2018hypothesis})
using a simple transformer baseline, finding that \textbf{single sentence relation leakage remains a severe problem}.


\paragraph{NLI models still use the leakage features to cheat.} We analyze the datasets using \textit{output decision agreement} and \textit{input token importance} statistics between models trained in the SSC and PSC to demonstrate this.


\paragraph{Automated leakage feature detection is feasible.} We introduce a novel model-based metric and dataset analysis tool, the \textit{Progressive Evaluation of Cluster Outliers} (PECO) (\autoref{fig:clust-snli-test}), for  {examining the degree of single sentence relation leakage} and eliminating it in future datasets\footnote{Code at 
\href{https://github.com/michaelsaxon/DatasetAnalysis}{\texttt{github.com/michaelsaxon/DatasetAnalysis}}, and an animated demo is available at \href{https://saxon.me/peco}{\texttt{saxon.me/peco}}.}. 




\begin{table*}[h!]
\footnotesize
    \centering
    \begin{tabular}{lllrrcrr}
         & & & \multicolumn{3}{|c|}{Accuracy (\% PSC)} & \multicolumn{2}{c|}{Hparams} \\
        \toprule
        Dataset & SOTA Model & Replication Model & SOTA & Ours & Maj. & LR & Batch \\
        \midrule
        SICK  &  NeuralLog \cite{chen-etal-2021-neurallog} &  \texttt{roberta-large} & 90.3 & 87.8 & 56.0 &  \lr{1}{5} & 18 \\
        SNLI  &  EFL \cite{wang2021entailment} &  \texttt{roberta-large} & 93.1 & 90.6 & 33.8 &  \lr{5}{6} & 128 \\
        MNLI-u  &  T5 \cite{raffel2019exploring} &  \texttt{roberta-large} & 92.0 & 88.7 & 35.6 &  \lr{5}{6} & 12 \\
        MNLI-b  &  \samel &  \samel  & 91.7 & 88.3 & 36.5 &  \lr{5}{6} & 12 \\
        XNLI &  ByT5 \cite{xue2021byt5} &  \texttt{xlm-roberta} & 83.7 & 73.7 & 33.3 &  \lr{1}{5} & 16 \\
        FEVER &  KILT \cite{petroni2021kilt} &  \texttt{roberta-large} & 86.3 & 74.7 & 33.3 &  \lr{1}{5} & 8 \\
        ANLI & InfoBERT \cite{wang2021infobert} &  \texttt{roberta-large} & 58.3 & 54.0 & 33.5 &  \lr{1}{6} & 18 \\
        \lastrow A1 & \samel &  \samel & 75.5 & 61.9 & 33.4 &  \lr{1}{6} & 18 \\
        \lastrow A2 & ALBERT \cite{Lan2020ALBERT} &  \samel & 58.6 & 50.0 & 33.4 &  \lr{1}{6} & 18 \\
        \lastrow A3 & \samel &  \samel & 53.4 & 49.8 & 33.5 &  \lr{1}{6} & 18 \\
        OCNLI & \texttt{RoBERTa-wwm-ext-l} \cite{xu2020clue} &  \texttt{bert-base-chinese} & 78.2 & 71.8 & 36.8 &  \lr{1}{5} &  128 \\
        CAugNLI & --- (Ours) &  \texttt{roberta-large} & 84.7 & 84.7 & 33.9 &  \lr{5}{6} & 64 \\
        SNLI$_\text{debiased}$ & \samel &  \texttt{roberta-large} & 95.6 & 95.6 & 35.5 &  \lr{1}{5} & 64 \\
        MNLI$_\text{debiased}$ & \samel &  \texttt{roberta-large} & 96.9 & 96.9 & 36.2 &  \lr{1}{5} & 16 \\        \bottomrule
    \end{tabular}
    \caption{For each dataset we analyze, the current state of the art (SOTA) model, base pretrained LM we use for replication, and accuracy for SOTA, our replication (Ours), and majority label-only (Maj.) classification in the standard paired-sentence NLI classification condition (PSC) with selected best-performing hyperparameters. 
    }
    \vspace{-2ex}
    \label{tab:training_details}
\end{table*}

\section{Quantifying NLI Dataset Bias}

An ideal NLI benchmark is neither ``saturated'' nor biased. 
Saturated benchmarks are datasets for which current approaches already achieve high accuracy.
They are ``solved'' and have limited utility in tracking future progress \cite{bowman2021will}.
We refer to as \textit{biased} any NLI benchmarks that exhibit significant single-sentence relation leakage through high achievable SSC accuracy in at least one single sentence condition.

We analyze 10 datasets containing a total of 14 test or validation sets in terms of \textit{biasedness} and \textit{saturatedness}, across 17 SSC conditions (premise-only ($s_1$) or hypothesis-only ($s_2$)).  \autoref{tab:dataset_info} provides an overview of this information along with statistics such as train/dev/test set size, and which sentence  population is potentially unbalanced. 

Each dataset $D$ is composed of $(s_1, s_2, r)$ tuples. We use the standard notation $(X_i, Y_i)\leftarrow D$ to describe the samples, as depending on whether the training condition is standard PSC or SSC, each $X_i$ can be $(s_{1i}, s_{2i})$, $s_{1i}$, or $s_{2i}$. Models trained in condition \textit{c} are referred to as $f_\text{c}$.

\subsection{Saturation and Bias Scoring}\label{subsec:comp}

We assess accuracy on the test set (or val set when no labeled test set is available) for each dataset in paired- and single-sentence conditions. 

\paragraph{Saturation (Accuracy):} We report state-of-the-art (SOTA) model performance results in the PSC,

\begin{equation} 
    \text{A}_\text{SOTA}(D) = P(f_\text{SOTA}(X_\text{test}) = Y_\text{test}); \:\:\:\: X,Y\in D
\end{equation}

For our model-level and sample-level comparative analysis between the PSC and SSC we train our own transformer-based models using a simple procedure (Sec. \ref{sec:repl}) to assess \textit{replication} accuracy,

\begin{equation}
    \text{A}_\text{PSC}(D) = P(f_\text{PSC}(X_\text{test}) = Y_\text{test}); \quad X,Y\in D
\end{equation}

\paragraph{SSC Accuracy:}

For each elicited population SSC we train a model $f_\text{SSC}$ according to the procedure described in Sec. \ref{sec:biastrain}, to assess $\text{A}_\text{SSC}$:

\begin{equation}
    \text{A}_\text{SSC}(D) = P(f_\text{SSC}(X_\text{test}) = Y_\text{test}); \quad X,Y\in D
\end{equation}

\begin{figure}[t!]
    \centering
    \includegraphics[width=0.9\linewidth]{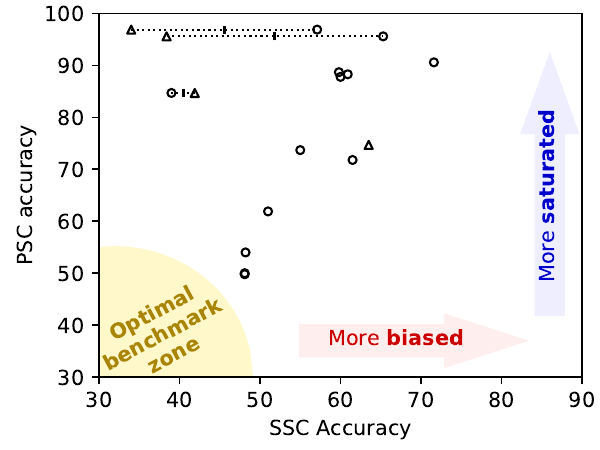}
    \caption{Where the existing NLI datasets fall on the SSC accuracy/PSC accuracy axis. An ideal benchmark isn't saturated (lower SOTA accuracy) and also is unbiased (low SSC accuracy). $\triangle$ markers represent premise-only SSC results and $\circ$ markers represent hypothesis-only SSCs. For datasets exhibiting leakage in both conditions, the two points are connected by a dotted line with a $|$ marker for their average SSC accuracy. 
    }
    \label{fig:bca_rep}
\end{figure}

\autoref{tab:training_details} shows current SOTA models and results for the 10 datasets, as well as our PSC model performance and the relevant training hyperparameters (more detail in Sec. \ref{sec:repl}).
\autoref{fig:bca_rep} shows  SSC accuracy against replication (PSC) accuracy for each dataset. 
Datasets that exhibit higher SSC accuracy have worse single sentence relation leakage, and are thereby questionable in their ability to capture reasoning abilities. 
Ideally, an \textbf{optimal benchmark} for NLI will both have low maximum SSC accuracy and low maximum PSC accuracy (room for future model growth).


These absolute measures of dataset bias and saturation are useful targets for future optimal benchmarks, it is important to understand how these measures interact with each other.

\subsection{Relative Dataset Bias Scoring}

We assess two relative dataset bias scores.

\paragraph{SSC Improvement over Chance:} We subtract the SSC test accuracy from the accuracy achieved by a ``guess majority label'' strategy following \cite{poliak2018hypothesis}:
\begin{equation}
\Delta_\text{maj} = P(f_\text{SSC}(X_\text{test})=Y_\text{test}) - P(Y_\text{maj}=Y_\text{test})
\end{equation}


This metric gives an insight into single sentence relation leakage that 
compensates for datasets (such as SICK) with an uneven base label distribution.

%


\paragraph{SSC-PSC Accuracy Recovered:} Accuracy achieved by SSC model over PSC:
\begin{equation}
\text{\%R}_\text{R} = \frac{P(f_\text{SSC}(X_\text{test})=Y_\text{test})}{ P(f_\text{PSC}(X_\text{test})=Y_\text{test})}
\end{equation}

This metric captures how similarly the single sentence and normal condition models perform.

\begin{table}[t!]
\small
    \centering
    \begin{tabular}{llrrr}
        \toprule
        Dataset & Cond. & SSC  & \%R$_\text{R}$ & $\Delta_{\text{maj}}$ \\
        \midrule
        SICK & $s_2$ & 60.0 & 68.3 & 4.0 \\
        SNLI & $s_2$ & 71.6 & 79.0 & 37.8 \\
        MNLI-b & $s_2$ & 59.8 & 67.4 & 24.2 \\
        MNLI-u & \samel & 60.9 & 69.0 & 24.4 \\
        XNLI & $s_2$ & 55.0 & 74.6 & 21.7 \\
        FEVER & $s_1$ & 63.5 & 85.0 & 30.2 \\
        ANLI & $s_2$ & 48.2 & 89.3 & 14.7 \\
        \lastrow A1 & \samel & 67.5 & 82.4 & 17.6 \\
        \lastrow A2 & \samel & 82.1 & 96.2 & 14.7 \\
        \lastrow A3 & \samel & 90.1 & 96.6 & 14.6 \\
        OCNLI & $s_2$ & 61.5 & 85.7 & 24.7 \\
        CAugNLI & $s_1$ & 41.9 & 49.5 & 8.0 \\
        \samel & $s_2$ & 39.0 & 46.0 & 5.1 \\
        SNLI$_\text{debiased}$ & $s_1$ & 45.3 & 45.3 & 2.9 \\
        \samel & $s_2$ & 65.3 & 68.3 & 29.8 \\
        MNLI$_\text{debiased}$ & $s_1$ & 34.0 & 35.6 & -2.2 \\
        \samel & $s_2$ & 57.1 & 58.9 & 20.9 \\           \bottomrule
        \end{tabular}
    \caption{
    For each NLI dataset and potential leakage-exhibiting single-sentence condition (Cond.) we report 
    \textit{test} accuracy in the single sentence condition (SSC), and three derived metrics from Sec. \ref{subsec:comp}.
    SOTA test accuracy recovery (\%R$_\text{S}$), 
    replication test recovery (\%R$_\text{R}$), and
    biased condition improvement over the chance majority guessing strategy ($\Delta_\text{maj}$).
    }
    \vspace{-2ex}
    \label{tab:results}
\end{table}

\subsection{Biased Model Results}
\autoref{tab:results} shows the extent of the single-sentence relation leakage problem across the 17 SSC tests on the 14 splits for the 10 NLI datasets.     
These results clearly show that 
\textit{each dataset exhibits significant single-sentence relation leakage} for at least one condition. 
The comparison columns replication test recovery (\%R$_\text{R}$), and SSC improvement over the chance majority guessing strategy ($\Delta_\text{maj}$) 
are all computed using the single sentence condition accuracy and the standard NLI two-sentence condition SOTA and replication accuracy values in \autoref{tab:training_details}.

\begin{figure}[t!]
    \centering
    \includegraphics[width=0.9\linewidth]{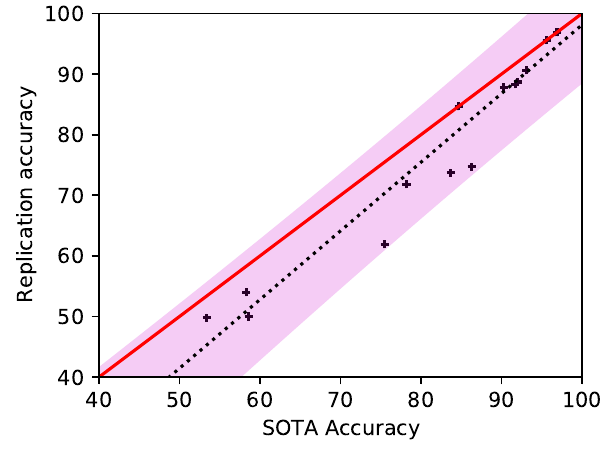}
    \caption{Test set accuracy for SOTA models vs our universal replication procedure models for each dataset in PSC, with a trendline (PCC=0.97) and  \textcolor{red}{$y=x$ line}. 
    }
    \label{fig:sota_rep}
    \vspace{-.5cm}
\end{figure}

\subsection{Model Training}\label{sec:repl}


Our training technique is simple and applied consistently to all datasets. 
We fine-tune three different language-specific pretrained transformer checkpoints\footnote{\texttt{roberta-large} \cite{liu2019roberta}, \texttt{xlm-roberta} \cite{conneau2019xlm}, \texttt{bert-base-chinese} \cite{devlin2018bert}} from \texttt{HuggingFace} \cite{wolf2019huggingface} using \texttt{Pytorch Lightning}. All models were trained on NVIDIA A-100 GPUs.
 All models are the HuggingFace \texttt{xForSequenceClassification} with \texttt{num\_classes=3} and no other modifications. All models are trained using the Adam optimizer with cross entropy loss. 

We find that this procedure produces broadly near-SOTA performance models, with a maximum relative accuracy difference of 8\%, and a 92\% Pearson's correlation coefficient (PCC) between SOTA and replication accuracy across the datasets (\autoref{fig:sota_rep}). \textcolor{Purple4}{\textbf{Our replications are a reasonable proxy to  SOTA for comparative dataset analysis}}.  


\subsubsection{Replication Training Details}\label{subsec:repl}

\paragraph{Hyperparameters} For each dataset in the paired sentence condition (PSC), we select a batch size for maximum GPU utilization. We perform a grid search over lr in \{\lr{5}{7}, \lr{1}{6}, \lr{5}{6}, \lr{1}{5}, \lr{5}{5}\}. 

\paragraph{Single Dataset Fine-tuning} We obtain separate fine-tuning checkpoints from the pretrained models for each dataset 
to enable clean analysis of one dataset at a time. We do not accumulate fine-tuning passes across multiple datasets.


\subsubsection{Single Sentence Condition Training}\label{sec:biastrain}


To train each dataset's corresponding  \textbf{SSC} model(s), we use the same setup as the PSC model but follow \citet{poliak2018hypothesis}'s formulation of fine-tuning the chosen classification model on only the SSC sentence,
\textit{premise only} or \textit{hypothesis only}. 

FEVER exhibits bias in the premise distribution, and CAugNLI, SNLI$_\text{debiased}$, and MNLI$_\text{debiased}$ exhibit imbalance in both (\autoref{tab:dataset_info}). For the datasets that have imbalanced distributions in both conditions, we separately train bias models for both hypothesis-only and premise-only. 

\section{Analyzing NLI Dataset Bias}\label{sec:comps}

In this section we introduce quantification techniques for more accurately characterizing the extent of these bias problems in the aforementioned NLI datasets, analyze how they interact with the observable bias itself, and develop tools for producing future NLI benchmarks that more closely resemble the ideal benchmark. 






\subsection{Sample-level Model Behavior}\label{sec:NBA}

We are particularly interested in understanding the degree to which models trained in the PSC and SSCs ``reason'' similarly. For this section we use the notations $f(\bm{X}_\text{test})$, $\bm{Y}_\text{test}$ to denote the ($1\times N$) column vectors of model output decisions and labels for a test set of $N$ samples, and  a simple agreement function $Ag(\bm{Y}_1, ..., \bm{Y}_n)$ 
as the ratio of elements that are identical across all $\bm{Y}_i$ to the vector size $N$. 
In other words, 
\begin{equation}
    Ag(f(\bm{X}_\text{test}), \bm{Y}_\text{test}) =P(Y_\text{test}=f(X_\text{test}))
\end{equation}

\paragraph{SSC-PSC Agreement (NBA):} The number of samples for which the SSC and PSC models agree over the total number of samples in the set:
\begin{equation}
\text{NBA} = \frac{Ag(f_\text{SSC}(\bm{X}_\text{test}),f_\text{PSC}(\bm{X}_\text{test}))}{|\bm{X}_\text{test}|}
\end{equation}

\paragraph{SSC-PSC Recovery (NBR):} The number of samples for which the SSC and PSC models agree, \textit{and both classify correctly} over the total number of samples they agree on:
\begin{equation}
\text{NBR} = \frac{Ag(f_\text{SSC}(\bm{X}_\text{test}),f_\text{PSC}(\bm{X}_\text{test}),\bm{Y}_\text{test})}{Ag(f_\text{SSC}(\bm{X}_\text{test}),f_\text{PSC}(\bm{X}_\text{test}))}
\end{equation}

\paragraph{Token Relevance Agreement (TRA):} 
Do SSC and PSC models reason alike? For a sentence $X$ with length $n$, we compute the gradient of the classification output posterior with respect to each token embedding $emb(w_j)$. We take the 2-norm of the each gradient vector and normalize it over the entire sequence to produce a normalized local explanation vector $m(f(X))$ \cite{sundararajan2017axiomatic}:

\begin{equation}
m(f, X)=\Big[\frac{|\nabla_{\text{emb}(w_j)}(f(\text{emb}(w_j)))|_2 }{\sum_{i=1}^n(|\nabla_{\text{emb}(w_i)}(f(\text{emb}(w_i)))|_2)}\Big]_{j=1}^n
\end{equation}

To compare ``reasoning'' similarity between the two models, we compute the samplewise input token relevance agreement can be computed using cosine similarity:

\begin{equation}
\text{TRA}(X_i) = \frac{m(f_\text{PSC}, X_i) \cdot m(f_\text{SSC}, X_i)}{
||m(f_\text{PSC}, X_i)||\, ||m(f_\text{SSC}, X_i)||}
\end{equation}

As the SSC and PSC inputs have different lengths, we pad the SSC importance vector for $m(f_\text{SSC}(X_i))$ with zeros either prepended or postpended (depending on if the SSC is hypothesis- or premise-only) to make the two local explanation map vectors of equal length. The dataset-level token relevance agreement is the average of samplewise TRA.

\subsection{Cluster-based Bias Evaluation}\label{sec:peco}


We are interested in investigating how the biased distributions of the elicited sentences in NLI datasets are captured in the learned representation spaces of models trained on them. In particular, we are interested in answering this question: \textbf{is elicited sentence label leakage captured semantically in regions of latent space}?

\begin{figure}[t!]
    \centering
    \includegraphics[width=1\linewidth]{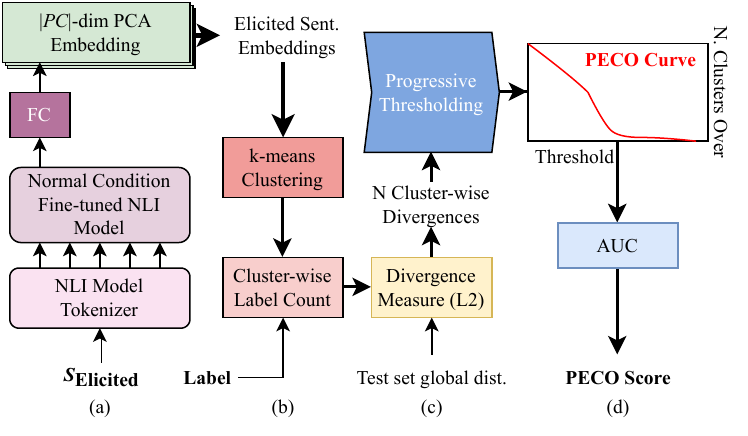}
    \caption{An overview of the approach to computing the PECO score from a collection of elicited population sentences $s_\text{elicited}$ 
    and their corresponding Labels $r$. When a fixed threshold is chosen, the Hypothesis embeddings can be dimensionality-reduced using T-SNE to produce plots like \autoref{fig:clust-snli-test}.
    }
    \label{fig:clusterbias}
    \vspace{-.3cm}
\end{figure}

To answer this we produce dimensionality-reduced \textit{elicited sentence embeddings} for the test set, using the PSC replication models, then fit a high-$k$ 
$k$-means clustering to this collection of embeddings. This will allow us to analyze how the local distribution of labels varies over the elicited sentence embedding space. By comparing the KL-divergence of the label distribution within each cluster and the global label distribution, we can compute the \textbf{Progressive Evaluation of Cluster Outliers} (PECO) score (\autoref{fig:clusterbias}).

\paragraph{Elicited Sentence Embeddings:} To embed the elicited sentences \textit{as they're learned by a model in the PSC}, we feed the elicited sentences $s_e$ through the PSC replication fine-tuned NLI model encoder. We extract the latent codes produced at the output very last fully connected layer of the model before the linear classifier to collect latent codes for every $s_e$ in the test set. We then embed these codes into their 30 principal components to produce the embeddings (\autoref{fig:clusterbias} (a)).

\paragraph{Clustering:} We fit a high-$k$ (in this case, $k=50$) $k$-means clustering over the distribution of elicited sentence embeddings to provide a set of local bins for analysis. For each cluster, we count the relation labels its samples contain, to produce a set of 50 cluster-label distributions (\autoref{fig:clusterbias} (b)).

\paragraph{Computing Cluster Divergences:} For each cluster label distribution $p_i = P(Y|\text{cluster}=i)$, we assess the L2 divergence between it and the global label distribution $p_\text{G}$ to produce divergence scores $s_i$: 
\begin{equation}
s_{i} = \frac{1}{3}\sum_{j=1}^3(P(Y=j) - P(Y=j|\text{cluster}=i))^2
\end{equation}
This step is depicted in \autoref{fig:clusterbias} (c).

\paragraph{Progressive Evaluation:} Finally, we compute the PECO score for this collection of cluster divergences as the area under the curve produced by counting the number clusters with divergence $s_i$ over some threshold $t$ for the range of $s_i$.
\begin{equation}
\text{PECO} = 100\int_{\min(s)}^{\max(s)}\frac{\text{count}_i(s_i>t)}{k}dt
\end{equation}

\paragraph{Generality of PECO:} These same techniques could be applied to  a wide variety of potential leakage features on the input to analyze a wide variety of correlation types. For example, input sentence words could be shuffled to test for word order invariance, or word classes could be specifically masked to test for spurious vocabulary correlations.


\paragraph{PECO Parameter Choices:} We discuss the impact of PECO parameters (e.g., choice of $k$-means, number of principal components to reduce to, use of L2 or KL-divergence) in \autoref{subsec:pecoparams}.


\begin{table}[t!]
\small
    \centering
    \begin{tabular}{llrrrr}
        \toprule
        Dataset & Cond.  & NBA & NBR & TRA & PECO \\
        \midrule
        SICK & $s_2$ & 48.8 & 49.7 & 63.6 & 13.9 \\
        SNLI & $s_2$ & 70.0 & 71.8 & 63.9 & 14 \\
        MNLI-b & $s_2$ & 32.5 & 33.4 & 57.6 & 6.5 \\
        MNLI-u & \samel & 47.5 & 48.6 & 59.3 & 7.5 \\
        XNLI & $s_2$ & 52.3 & 54.0 & 64.7 & 9 \\
        FEVER & $s_1$ & 39.4 & 38.0 & 50.7 & 14.3 \\
        ANLI & $s_2$ & 37.6 & 53.4 & 28.0 & 5.5 \\
        \lastrow A1 & \samel & 53.8 & 57.9 & 54.0 & 15.3 \\
        \lastrow A2 & \samel & 53.9 & 60.3 & 51.2 & 11.1 \\
        \lastrow A3 & \samel & 52.1 & 58.5 & 55.7 & 15.9 \\
        OCNLI & $s_2$ & 69.0 & 74.1 & 78.5 & 17.9 \\
        CAugNLI & $s_1$ & 32.9 & 32.7 & 39.5 & 5.5 \\
        \samel & $s_2$ & 43.4 & 42.5 & 59.7 & 5.4 \\
        SNLI$_\text{debiased}$ & $s_1$ & 32.0 & 31.8 & 41.5 & 8.2 \\
        \samel & $s_2$ & 60.9 & 61.4 & 58.4 & 8.3 \\
        MNLI$_\text{debiased}$ & $s_1$ & 34.7 & 34.0 & 46.7 & 6.6 \\
        \samel & $s_2$ & 42.2 & 42.0 & 58.0 & 6.8 \\         \bottomrule
        \end{tabular}
    \caption{Metrics comparing the behavior of our replication and single-sentence condition models on each dataset using the metrics introduced in Secs. \ref{sec:NBA}, \ref{sec:peco}:
    Normal-Bias Agreement (NBA) and Recovery (NBR), Token Relevance Agreement (TRA) and the Progressive Evaluation of Cluster Outliers (PECO) score. 
    }
    \label{tab:analysis}
\vspace{-.2cm}
\end{table}

\begin{figure}[b!]
\vspace{-.2cm}
    \centering
    \includegraphics[width=0.9\linewidth]{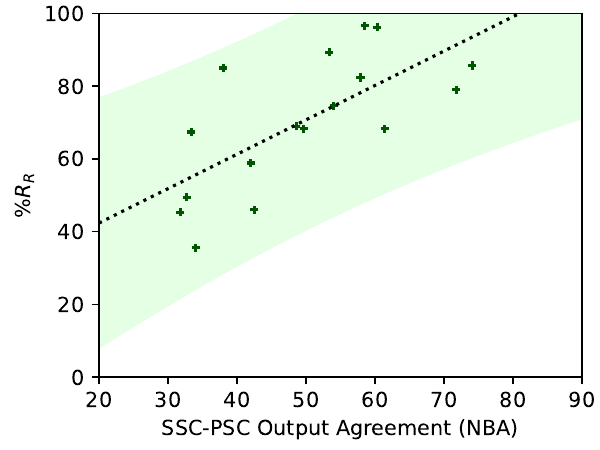}
    \caption{Model-wise output agreement vs Bias accuracy recovery (\%$R_R$). As the replication PSC model and the SSC model agree more often for a given dataset, their performances in the two conditions converge.
}
    \label{fig:outagr_recov}
\end{figure}

\section{Results}\label{sec:results}



As discussed above, ideal NLI benchmarks are neither \textit{saturated} nor \textit{biased}. Unfortunately, as \autoref{tab:results} demonstrates, none of the NLI datasets tested thus far satisfy this condition. This is more clearly illustrated in \autoref{fig:bca_rep}.
Two questions remain: \textit{to what extent do current models cheat} and \textit{how can we make less biased, less saturated datasets?} \autoref{tab:analysis} contains experimental results intended to answer these two questions.
The ``agreement metrics'' as introduced in Sec. \ref{sec:NBA}, SSC-PSC Agreement, SSC-PSC Recovery, and Token Relevance Agreement are provided in \autoref{tab:analysis}.



\subsection{Result-Metric Correlations}

We find that SSC-PSC model output agreement (NBA) and recovery rate \%R$_\text{R}$ are \textcolor{mdgreen}{\textbf{correlated with a \text{PCC} of 0.69}} (\autoref{fig:outagr_recov}). 
Datasets where the SSC and PSC models predict more similarly have closer the SSC and PSC performance results are for said datasets.
While this result is surprising, ANLI R3 is an interesting outlier (\autoref{sec:disc}).


\begin{figure}[t!]
    \centering
    \includegraphics[width=0.9\linewidth]{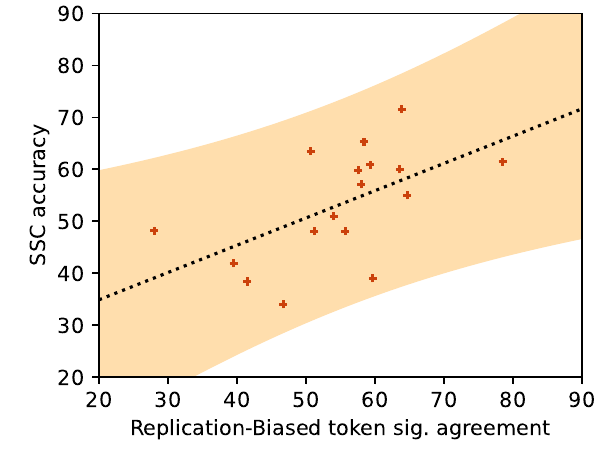}
    \caption{Token releveance agreement (TRA) vs SSC accuracy. When the PSC model and SSC models ``reason'' more similarly for a dataset, the relation leakage bias exhibited in that dataset tends to be higher.}
    \label{fig:tra_bca}
    \vspace{-.3cm}
\end{figure}

We find that TRA and SSC accuracy are also \textcolor{orange}{\textbf{positively correlated with a PCC of 0.57}} (\autoref{fig:tra_bca}). This result 
demonstrates that for a single dataset, similar reasoning patterns for the single sentence condition and standard sentence pair condition 
is strongly correlated to single-sentence relation leakage. In other words, \textbf{standard condition NLI models trained on biased (high leakage) datasets tend to cheat}. 
Thus, models indeed rely on annotation artifacts in NLI datasets to achieve high accuracy, and demonstrates that this continues to be a problem in newer NLI datasets, in spite of mitigation attempts. 
How can we use this knowledge to build better benchmarks?

\begin{figure}[t!]
    \centering
    \includegraphics[width=0.9\linewidth]{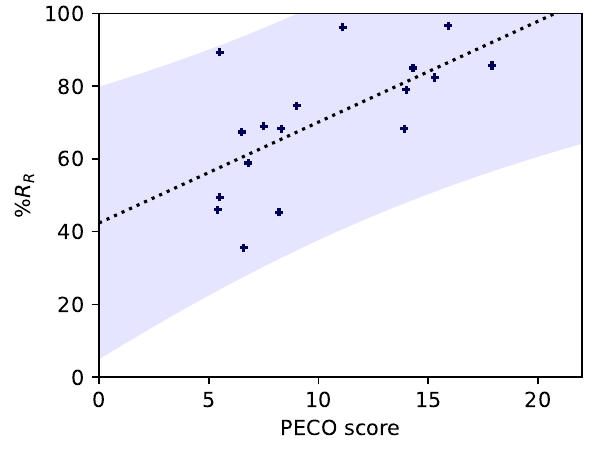}
    \caption{Test-set PECO score vs Bias accuracy recovery (\%$R_R$). This result suggests that interventions that produce lower PECO score datasets tend to yield datasets that exhibit less relation leakage in the SSC.
  }
    \label{fig:peco_rr}
    \vspace{-.2cm}
\end{figure}

\autoref{fig:peco_rr} depicts the relationship between PECO score and bias recovery (\%R$_\text{R}$). We find the two are \textcolor{darkblue}{\textbf{positively correlated with a PCC of 0.64}}. This result is fairly intuitive: the more uneven the distribution of labels is in the single-sentence latent spaces (and thus, the higher the area under the PECO curve), the more SSC performance approaches the standard PSC condition performance for a given NLI dataset. 
This suggests PECO-reducing interventions may be able to target debiasing efforts.

\section{Discussion}\label{sec:disc}



\paragraph{Some examples are only correctly classified in the single-sentence condition.}
A common assumption to discussions of cheating features in machine learning is that they play a role in inflated classification accuracy when present. However, ANLI R3 provides an interesting counterexample.
For this dataset, the hypothesis-only model achieves a SSC accuracy of 48.1\%, and the PSC model achieves 49.8\% (a \%R$_\text{R}$ of 90\%), and SOTA achieves 53\%. Despite this score similarity, the samples which the two conditions are able to actually classify correctly vary surprisingly. With an NBR of 58.5\%, only $\approx27\%$ of test samples are correctly classified by both the single and two sentence condition models. This means that around \textbf{21\% of samples in A3 test are \textit{only} correctly classified by the single-sentence model}.

Perhaps unsurprisingly, ANLI exhibits the lowest TRA out of all datasets tested, indicating that it is somewhat of an outlier in having the SSC and PSC condition models reason differently on it.



\paragraph{XNLI demonstrates the cross-lingual and semantic nature of single-sentence leakage.}
While previous work has focused on finding words, phrases, patterns, and heuristics in the surface form of the data, our study of XNLI provides an interesting opportunity to investigate the potential for the influence of underlying semantics as a leakage feature.
XNLI consists exclusively of 14 language test and val sets, manually translated from MNLI examples. Our XNLI PSC and SSC models are thus trained on MNLI alone, using the
multilingual \texttt{xlm-roberta} checkpoint.


This produces a natural experiment wherein surface form biases present in the training data are completely eradicated in the test set (as only the 14 non-English languages \autoref{tab:dataset_info} are present), while the underlying meanings encoded by those words remain.
In \autoref{tab:results} we indeed find that XNLI and MNLI exhibit very similar result comparisons. The models on both datasets have a $\Delta_\text{max}\approx20\%$ and $\%R_R\approx65$. These leakage feature results, being robust to manual translation into 14 different languages, seem to indicate that there is a strong fundamental semantic component to the human biases driving the elicited sentence relation leakage.

\paragraph{Relation leakage remains Unsolved.}
Elicited sentence relation leakage is a problem for all evaluated NLI datasets, including the new ones intended to fix it.
Recent datasets such as XNLI, FEVER, and OCNLI, 
exhibit high absolute SSC performance over majority ($\Delta_\text{maj}>20$).

Although ANLI 
and CAugNLI 
are improvements over the others in terms of $\Delta_\text{maj}$, with CAugNLI shining particularly in this regard, none eliminate the relation leakage problem entirely, as even CAugNLI still has $\Delta_\text{maj}=8.0$, an 8\% performance over chance in the single sentence condition.

SNLI$_\text{debiased}$ and MNLI$_\text{debiased}$, despite their intended purpose, still contain significant amounts of SS label leakage (29.8\% and 20.9\% over chance). This might be because while their production \cite{wu2022generating} does eliminate bias \textit{originally present in SNLI and MNLI}, it fails to prevent the introduction of \textit{new bias} in the data generation pipeline.


\paragraph{Cluster approaches are promising for future debiasing efforts.}
\autoref{fig:clust-snli-test} shows how the PECO-derived cluster-bias T-SNE plots can be used directly to visualize, analyze, and ``debug'' biased datasets. 
In the plot, SNLI clearly has considerably more high-biased clusters taking up a considerable portion of the latent space as compared to CAugNLI, for the bias threshold of 0.2.

An intervention could be performed on identified bias regions in the distribution by having human annotators create new premise sentences from the given hypotheses, thereby forcing the PECO-based bias metrics to reduce. This idea is further backed up by the PCC of 35.8 that we find between PECO and \%$\text{R}_\text{R}$, suggesting that producing datasets of lower PECO score will naturally lead to lower recovered performance in the SSC, and thus less elicited sentence relation leakage.






\section{Related Work}

\paragraph{Understanding Bias in NLI}
\citet{huang2020counterfactually} demonstrated that counterfactual augmentation alone cannot debias NLI.
Multi-task learning can improve model robustness to fitting spurious features \cite{tu2020empirical}, but 
because the underlying \textbf{benchmarks are biased}, progress on the desired reasoning capability is questionable \cite{poliak2018hypothesis}.
\citet{geva-etal-2019-modeling} strengthen the finding that annotator bias is a key driver of this poor generalization performance, showing that NLI models can struggle to even generalize across disjoint sets of annotators on the same task. 

Simple word- and n-gram level approaches have proven surprisingly capable in a-priori characterizations of dataset difficulty \cite{mckenna2020semantic} and producing difficult test sets \cite{saxon2021end} in diverse language domains such as SLU. \citet{gardner-etal-2021-competency} show how such purely frequentist approaches can identify word-level spurious correlations with respect to label class which drive in-part the shortcut features for classes of ``competency problems'' such as NLI.

\paragraph{Mitigating Bias in NLI}
\citet{belinkov-etal-2019-dont} demonstrate an approach to train NLI models robustly against some of these biases, using \citet{gururangan-2018-annotation}'s hard test set.
\citet{mccoy2020right} utilize simple heuristics like lexical overlap to produce the synthetic debiased HANS NLI dataset to test generalization.
This dataset has been used to evaluate techniques including predicate-argument- \cite{moosavi2020improving} and syntactic transformation-based \cite{min-etal-2020-syntactic} augmentations.
\citet{zhou-bansal-2020-towards} leverage a bag-of-words approach to debias datasets along lexical features. However, these approaches have yet to improve generalization in comprehensive replication studies has thus far \cite{bhargava-etal-2021-generalization}. 
Meanwhile \citet{varshney-etal-2022-unsupervised} propose a fully unsupervised data collection pipeline for NLI, in order to sidestep the problem of human biases entirely.

Approaches like \cite{kaushik2020learning} and \cite{wu2022generating} are very promising for producing data that reduces bias on a samplewise, but not populationwise level. Using our semantic, model-driven local bias finding strategies, future interventions can lead to the large scale production of debiased NLI datasets and a new generation of higher quality benchmarks for language understanding. \citet{Liu2022WANLIWA} perform such a targeted augmentation approach using the \textit{dataset cartography} sample characterization scheme from \citet{swayamdipta2020dataset} to produce WANLI, an NLI dataset that allows for improved performance on the aforementioned challenging test sets.

\subsection{PECO vs Dataset Cartography}

To determine if PECO-driven dataset augmentation is redundant given the recent release of WANLI, we seek to determine if usable samplewise information for targeting interventions (e.g., presence in a ``bias'' cluster) is captured during PECO analysis and is redundant to the relevant samplewise characterization produced in dataset cartography.

\begin{figure}[t!]
    \centering
    \includegraphics[width=1\linewidth]{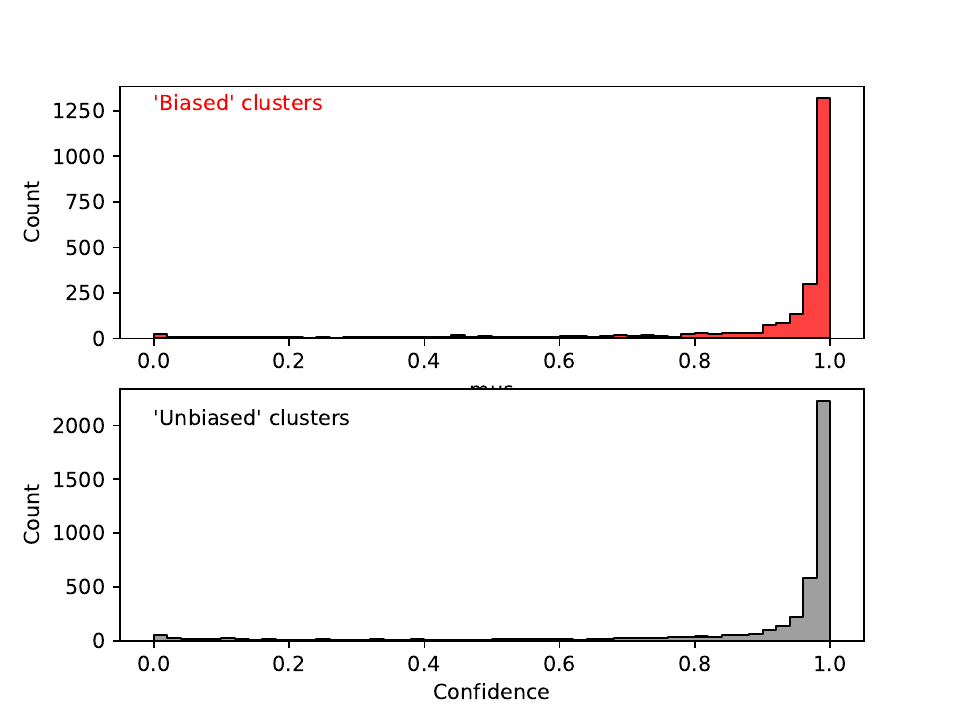}
    \caption{Histograms for the samplewise \textit{confidence} feature from dataset cartography in SNLI, for high- (biased) and low-$\text{PECO}_{L2}$ clusters.
  }
    \label{fig:peco_cartography}
    \vspace{-.2cm}
\end{figure}

To do this, we collect the samplewise \textit{confidence} feature \cite{swayamdipta2020dataset} during training of the PSC model for each validation set sample in SNLI. We then assign each validation sample to its corresponding PECO cluster (out of the 50) and produce two histograms of the confidence feature, one for \textbf{\textcolor{red}{``biased'' clusters}} ($\text{PECO}_{L2}>0.25$) and one for the other clusters. \autoref{fig:peco_cartography} shows the results of this experiment. We find that out of the 10k validation set examples, roughly 2 are assigned to an ``unbiased'' cluster for every 1 assigned to ``biased,'' roughly evenly across all confidence bins.

This result suggests that PECO clusterwise ``biasedness'' is orthogonal to the samplewise ease of learnability captured by the dataset cartography confidence feature. In other words, we find that \textbf{some samples are easy to learn (high confidence) because they are simple, while other samples are easy to learn because the model is using cheating features.} PECO-like analyses will be instrumental in guiding future efforts to eliminate shortcut features in natural language datasets.


\section{Conclusion}


In the half decade since \cite{poliak2018hypothesis} single sentence relation leakage bias has proven to remain a difficult issue. Efforts to debias NLI have led to datasets that merely exhibit different kinds of bias than those shown before, or less saturated benchmarks that continue to exhibit cheating features. Future work must prioritize reducing observable bias directly using a model-driven approach.


\section*{Limitations}

Our work is limited primarily by the PECO's reliance on test-set classification. To successfully analyze the train set-only datasets of SNLI$_\text{debiased}$ and MNLI$_\text{debiased}$, we had to generate our own train/test splits over the data by sampling. Luck in split selection may play a role in the level of observable bias in cases like these. Furthermore, this reliance on observing held-out samples to understand bias in general means that interventions to reduce single sentence label leakage must apply costly multi-fold splitting and analysis, consuming more significant compute resources than would otherwise be needed for other model-driven approaches.

\section*{Ethical Considerations}

In the short term, progress toward better natural language inference does not appear to lead to significant social risks in its broader impacts. While ``underclaiming'' progress in natural language processing tasks (e.g. exaggerating the scope or severity of failures of specific models on specific tasks) \cite{bowman2021combating} may be enabled by this work in the future, our focus on \textit{directly quantifiable and observable single sentence leakage}, use of SOTA-like models (fine-tuned transformers) for analysis, and our side-by-side comparison of our model implementations with SOTA all ensure that our criticisms of current NLI benchmarks are well-founded.
All data and tools we utilized were freely distributed for unlimited research use in the academic context.


\section*{Acknowledgements}

This work was supported in part by the National Science Foundation Graduate Research Fellowship under Grant No. 1650114. We would also like to thank the Robert N. Noyce Trust for their generous gift to University of California via the Noyce Initiative. The views and conclusions contained in this document are those of the authors and should not be interpreted as representing the sponsors.

\bibliography{clusterbias}

\begin{thebibliography}{50}
\expandafter\ifx\csname natexlab\endcsname\relax\def\natexlab#1{#1}\fi

\bibitem[{Agirre et~al.(2012)Agirre, Cer, Diab, and
  Gonzalez-Agirre}]{agirre-etal-2012-semeval}
Eneko Agirre, Daniel Cer, Mona Diab, and Aitor Gonzalez-Agirre. 2012.
\newblock \href {https://aclanthology.org/S12-1051} {{S}em{E}val-2012 task 6: A
  pilot on semantic textual similarity}.
\newblock In \emph{*{SEM} 2012: The First Joint Conference on Lexical and
  Computational Semantics {--} Volume 1: Proceedings of the main conference and
  the shared task, and Volume 2: Proceedings of the Sixth International
  Workshop on Semantic Evaluation ({S}em{E}val 2012)}, pages 385--393,
  Montr{\'e}al, Canada. Association for Computational Linguistics.

\bibitem[{Belinkov et~al.(2019)Belinkov, Poliak, Shieber, Van~Durme, and
  Rush}]{belinkov-etal-2019-dont}
Yonatan Belinkov, Adam Poliak, Stuart Shieber, Benjamin Van~Durme, and
  Alexander Rush. 2019.
\newblock \href {https://doi.org/10.18653/v1/P19-1084} {Don{'}t take the
  premise for granted: Mitigating artifacts in natural language inference}.
\newblock In \emph{Proceedings of the 57th Annual Meeting of the Association
  for Computational Linguistics}, pages 877--891, Florence, Italy. Association
  for Computational Linguistics.

\bibitem[{Bhargava et~al.(2021)Bhargava, Drozd, and
  Rogers}]{bhargava-etal-2021-generalization}
Prajjwal Bhargava, Aleksandr Drozd, and Anna Rogers. 2021.
\newblock \href {https://doi.org/10.18653/v1/2021.insights-1.18}
  {Generalization in {NLI}: Ways (not) to go beyond simple heuristics}.
\newblock In \emph{Proceedings of the Second Workshop on Insights from Negative
  Results in NLP}, pages 125--135, Online and Punta Cana, Dominican Republic.
  Association for Computational Linguistics.

\bibitem[{Bowman and Dahl(2021)}]{bowman2021will}
Samuel Bowman and George Dahl. 2021.
\newblock What will it take to fix benchmarking in natural language
  understanding?
\newblock In \emph{Proceedings of the 2021 Conference of the North American
  Chapter of the Association for Computational Linguistics: Human Language
  Technologies}, pages 4843--4855.

\bibitem[{Bowman(2021)}]{bowman2021combating}
Samuel~R Bowman. 2021.
\newblock When combating hype, proceed with caution.
\newblock \emph{arXiv preprint arXiv:2110.08300}.

\bibitem[{Bowman et~al.(2015)Bowman, Angeli, Potts, and
  Manning}]{bowman2015snli}
Samuel~R Bowman, Gabor Angeli, Christopher Potts, and Christopher~D Manning.
  2015.
\newblock The snli corpus.

\bibitem[{Chen et~al.(2021)Chen, Gao, and Moss}]{chen-etal-2021-neurallog}
Zeming Chen, Qiyue Gao, and Lawrence~S. Moss. 2021.
\newblock \href {https://doi.org/10.18653/v1/2021.starsem-1.7} {{N}eural{L}og:
  Natural language inference with joint neural and logical reasoning}.
\newblock In \emph{Proceedings of *SEM 2021: The Tenth Joint Conference on
  Lexical and Computational Semantics}, pages 78--88, Online. Association for
  Computational Linguistics.

\bibitem[{Conneau et~al.(2019)Conneau, Khandelwal, Goyal, Chaudhary, Wenzek,
  Guzm{\'{a}}n, Grave, Ott, Zettlemoyer, and Stoyanov}]{conneau2019xlm}
Alexis Conneau, Kartikay Khandelwal, Naman Goyal, Vishrav Chaudhary, Guillaume
  Wenzek, Francisco Guzm{\'{a}}n, Edouard Grave, Myle Ott, Luke Zettlemoyer,
  and Veselin Stoyanov. 2019.
\newblock \href {http://arxiv.org/abs/1911.02116} {Unsupervised cross-lingual
  representation learning at scale}.
\newblock \emph{CoRR}, abs/1911.02116.

\bibitem[{Conneau et~al.(2018)Conneau, Rinott, Lample, Williams, Bowman,
  Schwenk, and Stoyanov}]{conneau2018xnli}
Alexis Conneau, Ruty Rinott, Guillaume Lample, Adina Williams, Samuel~R.
  Bowman, Holger Schwenk, and Veselin Stoyanov. 2018.
\newblock Xnli: Evaluating cross-lingual sentence representations.
\newblock In \emph{Proceedings of the 2018 Conference on Empirical Methods in
  Natural Language Processing}. Association for Computational Linguistics.

\bibitem[{Devlin et~al.(2018)Devlin, Chang, Lee, and
  Toutanova}]{devlin2018bert}
Jacob Devlin, Ming-Wei Chang, Kenton Lee, and Kristina Toutanova. 2018.
\newblock Bert: Pre-training of deep bidirectional transformers for language
  understanding.
\newblock \emph{arXiv preprint arXiv:1810.04805}.

\bibitem[{Gardner et~al.(2021)Gardner, Merrill, Dodge, Peters, Ross, Singh, and
  Smith}]{gardner-etal-2021-competency}
Matt Gardner, William Merrill, Jesse Dodge, Matthew Peters, Alexis Ross, Sameer
  Singh, and Noah~A. Smith. 2021.
\newblock \href {https://doi.org/10.18653/v1/2021.emnlp-main.135} {Competency
  problems: On finding and removing artifacts in language data}.
\newblock In \emph{Proceedings of the 2021 Conference on Empirical Methods in
  Natural Language Processing}, pages 1801--1813, Online and Punta Cana,
  Dominican Republic. Association for Computational Linguistics.

\bibitem[{Geva et~al.(2019)Geva, Goldberg, and
  Berant}]{geva-etal-2019-modeling}
Mor Geva, Yoav Goldberg, and Jonathan Berant. 2019.
\newblock \href {https://doi.org/10.18653/v1/D19-1107} {Are we modeling the
  task or the annotator? an investigation of annotator bias in natural language
  understanding datasets}.
\newblock In \emph{Proceedings of the 2019 Conference on Empirical Methods in
  Natural Language Processing and the 9th International Joint Conference on
  Natural Language Processing (EMNLP-IJCNLP)}, pages 1161--1166, Hong Kong,
  China. Association for Computational Linguistics.

\bibitem[{Gururangan et~al.(2018)Gururangan, Swayamdipta, Levy, Schwartz,
  Bowman, and Smith}]{gururangan-2018-annotation}
Suchin Gururangan, Swabha Swayamdipta, Omer Levy, Roy Schwartz, Samuel Bowman,
  and Noah~A. Smith. 2018.
\newblock \href {https://doi.org/10.18653/v1/N18-2017} {Annotation artifacts in
  natural language inference data}.
\newblock In \emph{Proceedings of the 2018 Conference of the North {A}merican
  Chapter of the Association for Computational Linguistics: Human Language
  Technologies, Volume 2 (Short Papers)}, pages 107--112, New Orleans,
  Louisiana. Association for Computational Linguistics.

\bibitem[{Hu et~al.(2020)Hu, Richardson, Xu, Li, Kuebler, and Moss}]{ocnli}
Hai Hu, Kyle Richardson, Liang Xu, Lu~Li, Sandra Kuebler, and Larry Moss. 2020.
\newblock \href {https://arxiv.org/abs/2010.05444} {Ocnli: Original chinese
  natural language inference}.
\newblock In \emph{Findings of EMNLP}.

\bibitem[{Huang et~al.(2020)Huang, Liu, and Bowman}]{huang2020counterfactually}
William Huang, Haokun Liu, and Samuel~R Bowman. 2020.
\newblock Counterfactually-augmented snli training data does not yield better
  generalization than unaugmented data.
\newblock \emph{arXiv preprint arXiv:2010.04762}.

\bibitem[{Kaushik et~al.(2020)Kaushik, Setlur, Hovy, and
  Lipton}]{kaushik2020learning}
Divyansh Kaushik, Amrith Setlur, Eduard Hovy, and Zachary~C Lipton. 2020.
\newblock Explaining the efficacy of counterfactually augmented data.
\newblock \emph{International Conference on Learning Representations (ICLR)}.

\bibitem[{Krishna et~al.(2017)Krishna, Zhu, Groth, Johnson, Hata, Kravitz,
  Chen, Kalantidis, Li, Shamma, Bernstein, and
  Fei-Fei}]{10.1007/s11263-016-0981-7}
Ranjay Krishna, Yuke Zhu, Oliver Groth, Justin Johnson, Kenji Hata, Joshua
  Kravitz, Stephanie Chen, Yannis Kalantidis, Li-Jia Li, David~A. Shamma,
  Michael~S. Bernstein, and Li~Fei-Fei. 2017.
\newblock \href {https://doi.org/10.1007/s11263-016-0981-7} {Visual genome:
  Connecting language and vision using crowdsourced dense image annotations}.
\newblock \emph{Int. J. Comput. Vision}, 123(1):32–73.

\bibitem[{Lan et~al.(2020)Lan, Chen, Goodman, Gimpel, Sharma, and
  Soricut}]{Lan2020ALBERT}
Zhenzhong Lan, Mingda Chen, Sebastian Goodman, Kevin Gimpel, Piyush Sharma, and
  Radu Soricut. 2020.
\newblock \href {https://openreview.net/forum?id=H1eA7AEtvS} {Albert: A lite
  bert for self-supervised learning of language representations}.
\newblock In \emph{International Conference on Learning Representations}.

\bibitem[{Liu et~al.(2022)Liu, Swayamdipta, Smith, and Choi}]{Liu2022WANLIWA}
Alisa Liu, Swabha Swayamdipta, Noah~A. Smith, and Yejin Choi. 2022.
\newblock Wanli: Worker and ai collaboration for natural language inference
  dataset creation.
\newblock In \emph{Conference on Empirical Methods in Natural Language
  Processing}.

\bibitem[{Liu et~al.(2019)Liu, Ott, Goyal, Du, Joshi, Chen, Levy, Lewis,
  Zettlemoyer, and Stoyanov}]{liu2019roberta}
Yinhan Liu, Myle Ott, Naman Goyal, Jingfei Du, Mandar Joshi, Danqi Chen, Omer
  Levy, Mike Lewis, Luke Zettlemoyer, and Veselin Stoyanov. 2019.
\newblock Roberta: A robustly optimized bert pretraining approach.
\newblock \emph{arXiv preprint arXiv:1907.11692}.

\bibitem[{Maas et~al.(2011)Maas, Daly, Pham, Huang, Ng, and
  Potts}]{maas-etal-2011-learning}
Andrew~L. Maas, Raymond~E. Daly, Peter~T. Pham, Dan Huang, Andrew~Y. Ng, and
  Christopher Potts. 2011.
\newblock \href {https://aclanthology.org/P11-1015} {Learning word vectors for
  sentiment analysis}.
\newblock In \emph{Proceedings of the 49th Annual Meeting of the Association
  for Computational Linguistics: Human Language Technologies}, pages 142--150,
  Portland, Oregon, USA. Association for Computational Linguistics.

\bibitem[{Marelli et~al.(2014)Marelli, Menini, Baroni, Bentivogli, Bernardi,
  and Zamparelli}]{marelli2014sick}
Marco Marelli, Stefano Menini, Marco Baroni, Luisa Bentivogli, Raffaella
  Bernardi, and Roberto Zamparelli. 2014.
\newblock A sick cure for the evaluation of compositional distributional
  semantic models.
\newblock In \emph{Proceedings of the Ninth International Conference on
  Language Resources and Evaluation (LREC'14)}, pages 216--223.

\bibitem[{McCoy et~al.(2020)McCoy, Pavlick, and Linzen}]{mccoy2020right}
R~Thomas McCoy, Ellie Pavlick, and Tal Linzen. 2020.
\newblock Right for the wrong reasons: Diagnosing syntactic heuristics in
  natural language inference.
\newblock In \emph{57th Annual Meeting of the Association for Computational
  Linguistics, ACL 2019}, pages 3428--3448. Association for Computational
  Linguistics (ACL).

\bibitem[{McKenna et~al.(2020)McKenna, Choudhary, Saxon, Strimel, and
  Mouchtaris}]{mckenna2020semantic}
Joseph~P. McKenna, Samridhi Choudhary, Michael Saxon, Grant~P. Strimel, and
  Athanasios Mouchtaris. 2020.
\newblock \href {https://doi.org/10.21437/Interspeech.2020-2929} {{Semantic
  Complexity in End-to-End Spoken Language Understanding}}.
\newblock In \emph{Proc. Interspeech 2020}, pages 4273--4277.

\bibitem[{Min et~al.(2020)Min, McCoy, Das, Pitler, and
  Linzen}]{min-etal-2020-syntactic}
Junghyun Min, R.~Thomas McCoy, Dipanjan Das, Emily Pitler, and Tal Linzen.
  2020.
\newblock \href {https://doi.org/10.18653/v1/2020.acl-main.212} {Syntactic data
  augmentation increases robustness to inference heuristics}.
\newblock In \emph{Proceedings of the 58th Annual Meeting of the Association
  for Computational Linguistics}, pages 2339--2352, Online. Association for
  Computational Linguistics.

\bibitem[{Moosavi et~al.(2020)Moosavi, de~Boer, Utama, and
  Gurevych}]{moosavi2020improving}
Nafise~Sadat Moosavi, Marcel de~Boer, Prasetya~Ajie Utama, and Iryna Gurevych.
  2020.
\newblock Improving robustness by augmenting training sentences with
  predicate-argument structures.
\newblock \emph{arXiv preprint arXiv:2010.12510}.

\bibitem[{Nie et~al.(2019)Nie, Chen, and Bansal}]{nie2019combining}
Yixin Nie, Haonan Chen, and Mohit Bansal. 2019.
\newblock Combining fact extraction and verification with neural semantic
  matching networks.
\newblock In \emph{Association for the Advancement of Artificial Intelligence
  ({AAAI})}.

\bibitem[{Nie et~al.(2020)Nie, Williams, Dinan, Bansal, Weston, and
  Kiela}]{nie2020adversarial}
Yixin Nie, Adina Williams, Emily Dinan, Mohit Bansal, Jason Weston, and Douwe
  Kiela. 2020.
\newblock \href {http://arxiv.org/abs/1910.14599} {Adversarial {NLI}: {A} {New}
  {Benchmark} for {Natural} {Language} {Understanding}}.
\newblock \emph{arXiv:1910.14599 [cs]}.
\newblock ArXiv: 1910.14599.

\bibitem[{Petroni et~al.(2021)Petroni, Piktus, Fan, Lewis, Yazdani, De~Cao,
  Thorne, Jernite, Karpukhin, Maillard, Plachouras, Rockt{\"a}schel, and
  Riedel}]{petroni2021kilt}
Fabio Petroni, Aleksandra Piktus, Angela Fan, Patrick Lewis, Majid Yazdani,
  Nicola De~Cao, James Thorne, Yacine Jernite, Vladimir Karpukhin, Jean
  Maillard, Vassilis Plachouras, Tim Rockt{\"a}schel, and Sebastian Riedel.
  2021.
\newblock \href {https://doi.org/10.18653/v1/2021.naacl-main.200} {{KILT}: a
  benchmark for knowledge intensive language tasks}.
\newblock In \emph{Proceedings of the 2021 Conference of the North American
  Chapter of the Association for Computational Linguistics: Human Language
  Technologies}, pages 2523--2544, Online. Association for Computational
  Linguistics.

\bibitem[{Poliak et~al.(2018)Poliak, Naradowsky, Haldar, Rudinger, and
  Van~Durme}]{poliak2018hypothesis}
Adam Poliak, Jason Naradowsky, Aparajita Haldar, Rachel Rudinger, and Benjamin
  Van~Durme. 2018.
\newblock Hypothesis only baselines in natural language inference.
\newblock In \emph{Proceedings of the Seventh Joint Conference on Lexical and
  Computational Semantics}, pages 180--191.

\bibitem[{Radford et~al.(2019)Radford, Wu, Child, Luan, Amodei, Sutskever
  et~al.}]{radford2019language}
Alec Radford, Jeffrey Wu, Rewon Child, David Luan, Dario Amodei, Ilya
  Sutskever, et~al. 2019.
\newblock Language models are unsupervised multitask learners.
\newblock \emph{OpenAI blog}, 1(8):9.

\bibitem[{Raffel et~al.(2019)Raffel, Shazeer, Roberts, Lee, Narang, Matena,
  Zhou, Li, and Liu}]{raffel2019exploring}
Colin Raffel, Noam Shazeer, Adam Roberts, Katherine Lee, Sharan Narang, Michael
  Matena, Yanqi Zhou, Wei Li, and Peter~J Liu. 2019.
\newblock Exploring the limits of transfer learning with a unified text-to-text
  transformer.
\newblock \emph{arXiv preprint arXiv:1910.10683}.

\bibitem[{Saxon et~al.(2021)Saxon, Choudhary, McKenna, and
  Mouchtaris}]{saxon2021end}
Michael Saxon, Samridhi Choudhary, Joseph~P. McKenna, and Athanasios
  Mouchtaris. 2021.
\newblock \href {https://doi.org/10.21437/Interspeech.2021-1826} {{End-to-End
  Spoken Language Understanding for Generalized Voice Assistants}}.
\newblock In \emph{Proc. Interspeech 2021}, pages 4738--4742.

\bibitem[{Sundararajan et~al.(2017)Sundararajan, Taly, and
  Yan}]{sundararajan2017axiomatic}
Mukund Sundararajan, Ankur Taly, and Qiqi Yan. 2017.
\newblock Axiomatic attribution for deep networks.
\newblock In \emph{International conference on machine learning}, pages
  3319--3328. PMLR.

\bibitem[{Swayamdipta et~al.(2020)Swayamdipta, Schwartz, Lourie, Wang,
  Hajishirzi, Smith, and Choi}]{swayamdipta2020dataset}
Swabha Swayamdipta, Roy Schwartz, Nicholas Lourie, Yizhong Wang, Hannaneh
  Hajishirzi, Noah~A Smith, and Yejin Choi. 2020.
\newblock Dataset cartography: Mapping and diagnosing datasets with training
  dynamics.
\newblock In \emph{Proceedings of the 2020 Conference on Empirical Methods in
  Natural Language Processing (EMNLP)}, pages 9275--9293.

\bibitem[{Thorne et~al.(2018)Thorne, Vlachos, Christodoulopoulos, and
  Mittal}]{thorne2018fever}
James Thorne, Andreas Vlachos, Christos Christodoulopoulos, and Arpit Mittal.
  2018.
\newblock Fever: a large-scale dataset for fact extraction and verification.
\newblock In \emph{NAACL-HLT}.

\bibitem[{Tu et~al.(2020)Tu, Lalwani, Gella, and He}]{tu2020empirical}
Lifu Tu, Garima Lalwani, Spandana Gella, and He~He. 2020.
\newblock \href {https://doi.org/10.1162/tacl_a_00335} {{An Empirical Study on
  Robustness to Spurious Correlations using Pre-trained Language Models}}.
\newblock \emph{Transactions of the Association for Computational Linguistics},
  8:621--633.

\bibitem[{Varshney et~al.(2022)Varshney, Banerjee, Gokhale, and
  Baral}]{varshney-etal-2022-unsupervised}
Neeraj Varshney, Pratyay Banerjee, Tejas Gokhale, and Chitta Baral. 2022.
\newblock \href {https://doi.org/10.18653/v1/2022.findings-acl.159}
  {Unsupervised natural language inference using {PHL} triplet generation}.
\newblock In \emph{Findings of the Association for Computational Linguistics:
  ACL 2022}, pages 2003--2016, Dublin, Ireland. Association for Computational
  Linguistics.

\bibitem[{Wang et~al.(2021{\natexlab{a}})Wang, Wang, Cheng, Gan, Jia, Li, and
  Liu}]{wang2021infobert}
Boxin Wang, Shuohang Wang, Yu~Cheng, Zhe Gan, Ruoxi Jia, Bo~Li, and Jingjing
  Liu. 2021{\natexlab{a}}.
\newblock \href {https://openreview.net/forum?id=hpH98mK5Puk}
  {Info{\{}bert{\}}: Improving robustness of language models from an
  information theoretic perspective}.
\newblock In \emph{International Conference on Learning Representations}.

\bibitem[{Wang et~al.(2021{\natexlab{b}})Wang, Fang, Khabsa, Mao, and
  Ma}]{wang2021entailment}
Sinong Wang, Han Fang, Madian Khabsa, Hanzi Mao, and Hao Ma.
  2021{\natexlab{b}}.
\newblock Entailment as few-shot learner.
\newblock \emph{arXiv preprint arXiv:2104.14690}.

\bibitem[{Wang et~al.(2021{\natexlab{c}})Wang, Chen, Saxon, and
  Wang}]{wang2021counterfactual}
Xinyi Wang, Wenhu Chen, Michael Saxon, and William~Yang Wang.
  2021{\natexlab{c}}.
\newblock Counterfactual maximum likelihood estimation for training deep
  networks.
\newblock \emph{arXiv preprint arXiv:2106.03831}.

\bibitem[{Williams et~al.(2018)Williams, Nangia, and Bowman}]{williams2018mnli}
Adina Williams, Nikita Nangia, and Samuel Bowman. 2018.
\newblock \href {http://aclweb.org/anthology/N18-1101} {A broad-coverage
  challenge corpus for sentence understanding through inference}.
\newblock In \emph{Proceedings of the 2018 Conference of the North American
  Chapter of the Association for Computational Linguistics: Human Language
  Technologies, Volume 1 (Long Papers)}, pages 1112--1122. Association for
  Computational Linguistics.

\bibitem[{Wolf et~al.(2019)Wolf, Debut, Sanh, Chaumond, Delangue, Moi, Cistac,
  Rault, Louf, Funtowicz et~al.}]{wolf2019huggingface}
Thomas Wolf, Lysandre Debut, Victor Sanh, Julien Chaumond, Clement Delangue,
  Anthony Moi, Pierric Cistac, Tim Rault, R{\'e}mi Louf, Morgan Funtowicz,
  et~al. 2019.
\newblock Huggingface's transformers: State-of-the-art natural language
  processing.
\newblock \emph{arXiv preprint arXiv:1910.03771}.

\bibitem[{Wu et~al.(2022)Wu, Gardner, Stenetorp, and Dasigi}]{wu2022generating}
Yuxiang Wu, Matt Gardner, Pontus Stenetorp, and Pradeep Dasigi. 2022.
\newblock Generating data to mitigate spurious correlations in natural language
  inference datasets.
\newblock In \emph{Proceedings of the 60th Annual Meeting of the Association
  for Computational Linguistics}. Association for Computational Linguistics.

\bibitem[{Xu et~al.(2020)Xu, Hu, Zhang, Li, Cao, Li, Xu, Sun, Yu, Yu
  et~al.}]{xu2020clue}
Liang Xu, Hai Hu, Xuanwei Zhang, Lu~Li, Chenjie Cao, Yudong Li, Yechen Xu, Kai
  Sun, Dian Yu, Cong Yu, et~al. 2020.
\newblock Clue: A chinese language understanding evaluation benchmark.
\newblock In \emph{Proceedings of the 28th International Conference on
  Computational Linguistics}, pages 4762--4772.

\bibitem[{Xue et~al.(2021)Xue, Barua, Constant, Al-Rfou, Narang, Kale, Roberts,
  and Raffel}]{xue2021byt5}
Linting Xue, Aditya Barua, Noah Constant, Rami Al-Rfou, Sharan Narang, Mihir
  Kale, Adam Roberts, and Colin Raffel. 2021.
\newblock \href {https://arxiv.org/abs/2105.13626} {Byt5: Towards a token-free
  future with pre-trained byte-to-byte models}.
\newblock \emph{CoRR}, abs/2105.13626.

\bibitem[{Yang et~al.(2018)Yang, Qi, Zhang, Bengio, Cohen, Salakhutdinov, and
  Manning}]{yang-etal-2018-hotpotqa}
Zhilin Yang, Peng Qi, Saizheng Zhang, Yoshua Bengio, William Cohen, Ruslan
  Salakhutdinov, and Christopher~D. Manning. 2018.
\newblock \href {https://doi.org/10.18653/v1/D18-1259} {{H}otpot{QA}: A dataset
  for diverse, explainable multi-hop question answering}.
\newblock In \emph{Proceedings of the 2018 Conference on Empirical Methods in
  Natural Language Processing}, pages 2369--2380, Brussels, Belgium.
  Association for Computational Linguistics.

\bibitem[{Young et~al.(2014)Young, Lai, Hodosh, and
  Hockenmaier}]{young-etal-2014-image}
Peter Young, Alice Lai, Micah Hodosh, and Julia Hockenmaier. 2014.
\newblock \href {https://doi.org/10.1162/tacl_a_00166} {From image descriptions
  to visual denotations: New similarity metrics for semantic inference over
  event descriptions}.
\newblock \emph{Transactions of the Association for Computational Linguistics},
  2:67--78.

\bibitem[{Zhang et~al.(2019)Zhang, Bai, Liang, Bai, Chang, Yu, Zhu, and
  Zhao}]{zhang_selection_2019}
Guanhua Zhang, Bing Bai, Jian Liang, Kun Bai, Shiyu Chang, Mo~Yu, Conghui Zhu,
  and Tiejun Zhao. 2019.
\newblock \href {https://doi.org/10.18653/v1/P19-1435} {Selection {Bias}
  {Explorations} and {Debias} {Methods} for {Natural} {Language} {Sentence}
  {Matching} {Datasets}}.
\newblock In \emph{Proceedings of the 57th {Annual} {Meeting} of the
  {Association} for {Computational} {Linguistics}}, pages 4418--4429, Florence,
  Italy. Association for Computational Linguistics.

\bibitem[{Zhou and Bansal(2020)}]{zhou-bansal-2020-towards}
Xiang Zhou and Mohit Bansal. 2020.
\newblock \href {https://doi.org/10.18653/v1/2020.acl-main.773} {Towards
  robustifying {NLI} models against lexical dataset biases}.
\newblock In \emph{Proceedings of the 58th Annual Meeting of the Association
  for Computational Linguistics}, pages 8759--8771, Online. Association for
  Computational Linguistics.

\end{thebibliography}
\bibliographystyle{acl_natbib}

\appendix

\section{Detailed Dataset Info}

\paragraph{SICK} Sentences Involving Compositional Knowledge \cite{marelli2014sick} was produced by instructing annotators to label existing sourced pairs from 8K ImageFlickr data set \cite{young-etal-2014-image} and SemEval 2012 STS MSR-Video Description data set \cite{agirre-etal-2012-semeval}. The dataset is in English. Each sentence pair was annotated for relatedness and entailment by means of crowdsourcing techniques.

\paragraph{SNLI} The Stanford NLI dataset was produced using \cite{bowman2015snli}. The corpus contains content from the Flickr 30k Corpus \cite{young-etal-2014-image}, VisualGenome corpus \cite{10.1007/s11263-016-0981-7} and \citet{gururangan-2018-annotation}. The corpus is in English. The dataset is collected through human-written English sentence pairs.

\paragraph{MNLI} The Multi-genre NLI Corpus \cite{williams2018mnli} is modeled on the SNLI corpus \cite{bowman2015snli} but it differs in the range of genres of spoken and written English text supporting cross-genre evaluation.

\paragraph{XNLI} The Cross-Lingual NLI Corpus \cite{conneau2018xnli} consists of manually-translated dev and test samples from MNLI in 14 languages: French, Spanish, German, Greek, Belgian, Russian, Turkish, Arabic, Vietnamese, Thai, Chinese, Hindi, Swahili, and Urdu. 
It is interesting for analysis because on a high level the semantics of the data follow MNLI. The corpus is made to evaluate the inference in any language when only English data is presented at training time.

\paragraph{FEVER} NLI-style FEVER \cite{nie2019combining} is an NLI reformulation of the FEVER claim verification dataset \cite{thorne2018fever}. The original dataset was collected by eliciting annotators to write fact sentences that are \textit{supported}, \textit{refuted}, or \textit{unverifiable} relative source passages drawn from Wikipedia. This is converted into an NLI task by treating the elicited sentences as premises and the source passages as NLI pairs with relations \textit{entail}, \textit{contradict}, or \textit{neutral} respectively. This dataset is unique in that the premises were elicited from seed hypotheses, meaning it has a balanced hyp. distribution but potentially biased prem. distribution.

\paragraph{ANLI} The adversarial NLI corpus \cite{nie2020adversarial} is collected through crowdworkers and the purpose of this dataset creation is to make the state-of-art results fail in this dataset. The sentences are selected from the Wekipedia and manually curated HotpotQA training set \cite{yang-etal-2018-hotpotqa}. The language is in English. It contains three partitions of increasing complexity and size, which we refer to hereafter as A1, A2, and A3. Detailed data statistics are in Table \ref{tab:dataset_info}.

\paragraph{OCNLI} The Original Chinese NLI corpus was collected following MNLI-procedures but with strategies intended to produce challenging inference pairs \cite{ocnli}. No translation was employed in producing this data; the source premise sentences and elicited hypotheses are original.

\paragraph{CAugNLI} \citet{kaushik2020learning} produced counterfactually augmented datasets for NLI and sentiment analysis using human annotators, instructing them to make minimal changes to the sentences beyond those necessary to change the label. It extends the work of \citet{maas-etal-2011-learning} and \citet{bowman2015snli}. They find that a BiLSTM classifier achieves negligible performance over chance when trained on hypothesis only. However, since their dataset includes elicited modified sentences in both the premise and hypothesis populations, there are opportunities for bias on both.

CAugNLI was produced by having human annotators minimally modify either the premise or hypothesis of 2,500 samples drawn randomly from SNLI so as to produce new samples with similar structure and word distributions but different meanings. 
These modifications are intended to reduce spurious correlations, in particular by roughly equalizing the distribution of relation labels with respect to word-level and semantic-level patterns in the elicited hypothesis sentences.

\paragraph{SNLI$_\text{debiased}$ and MNLI$_\text{db}$} are augmentations of the SNLI and MNLI train sets produced by training GPT-2 \cite{radford2019language} generators on them, and then generating samples which they check for accuracy using a pretrained RoBERTa NLI classifier, and then reject if they exhibit spurious correlations including
samplewise hypothesis-only model classifiability \cite{wu2022generating}. To do this they first train static hypothesis-only SNLI and MNLI models, and reject all generated samples that can be successfully classified hypothesis-only by them. However, beyond this test under a static hypothesis-only distribution they do not attempt to assess if their generator models introduce \textit{new leakage features} in the sentence distributions as a result of their accuracy filtering process. To test this we create test splits on the data (as they provide train sets only) which contain no sentence overlap with the train sets through random sampling.

\section{Training on PSC, Testing on SSC}

Here we justify why PECO is computed on single-sentence condition (SSC) examples, using \textit{models trained on the paired-sentence condition} (PSC).

Our core goal is to characterize only the \textbf{model-relevant} shortcut features that are present in the SSC data, to enable better model-level understanding and to enable shortcut feature elimination in future datasets.
While all SSC accuracy must be driven by SSC-visible shortcuts, it is possible that \textbf{some SSC-visible cheating features aren't actually used as shortcuts by PSC classifiers}.
Thus, we have to train on PSC and test on SSC, and PECO is an alternative metric of bias that captures this model-level separability of sentences in the SSC notion better than other approaches.

\section{PECO Parameter Details}\label{subsec:pecoparams}

The PECO scoring pipeline contains a number of parameters that require motivation, including SSC and PSC model training hyperparameters, number of principal components to reduce to during PCA $|PC|$, and number of $k$-means clusters $k$ to divide the test set into for analysis (\autoref{fig:clusterbias}). We specify our NN hyperparams that we performed grid search over in \autoref{subsec:repl}. However, selecting $k$ and $|PC|$ is not a straightforward simple grid search.

We report PECO scores for all assessed datasets in \autoref{tab:full_hparam_1}, for $k\in{10,25,50,100}$ and no PCA projection, $|PC|=50$, and $|PC|=100$ PCA conditions. We find that for a given $k$, the different PCA conditions have limited impact on the final scores. We also find that L2- and KLD-based PECO scores are well-correlated. For smaller test sets (e.g., ANLI and its partitions A1-A3, OCNLI) there is increased sensitivity to variations in $k$ relative to the larger datasets such as SNLI. We selected $k=30$, $|PC|=50$ for our main experimental PECO results as it didn't produce the extreme swings in score for small test sets that we observed for higher $k$.

\begin{table*}[t!]
\small
    \centering
    \begin{tabular}{llrrr rrr rrr rrr}
        \toprule
                &       &     \multicolumn{6}{c}{\textit{k=10}} & \multicolumn{6}{c}{\textit{k=25}}\\
                &       &      \multicolumn{2}{c}{No PCA} & \multicolumn{2}{c}{$|PC|=50$} & \multicolumn{2}{c}{$|PC|=100$} & \multicolumn{2}{c}{No PCA} & \multicolumn{2}{c}{$|PC|=50$} & \multicolumn{2}{c}{$|PC|=100$} \\
        Dataset & Cond.  & L2 & KLD & L2 & KLD & L2 & KLD & L2 & KLD & L2 & KLD & L2 & KLD \\
        \midrule
        SICK & $s_2$ & 6.0 & 0.085 & 6.0 & 0.075 & 6.0 & 0.080 & 6.8 & 0.084 & 6.0 & 0.080 & 6.0 & 0.078 \\
        SNLI & $s_2$ & 8.5 & 0.155 & 8.5 & 0.150 & 8.5 & 0.150 & 8.4 & 0.138 & 8.8 & 0.140 & 8.4 & 0.136 \\
        MNLI-b & $s_2$ & 5.0 & 0.055 & 5.0 & 0.055 & 5.0 & 0.055 & 5.2 & 0.064 & 5.4 & 0.064 & 5.2 & 0.060 \\
        MNLI-u & \samel & 5.5 & 0.065 & 5.5 & 0.065 & 5.5 & 0.065 & 5.8 & 0.072 & 5.6 & 0.072 & 5.8 & 0.070 \\
        XNLI & $s_2$ & 5.0 & 0.065 & 5.0 & 0.065 & 5.0 & 0.065 & 5.2 & 0.068 & 5.2 & 0.068 & 5.2 & 0.068 \\
        FEVER & $s_1$ & 11.0 & 0.160 & 11.0 & 0.160 & 11.0 & 0.160 & 13.2 & 0.154 & 10.6 & 0.150 & 10.2 & 0.144 \\
        ANLI & $s_2$ & 15.0 & 0.095 & 14.5 & 0.070 & 14.5 & 0.070 & 31.0 & 0.176 & 8.8 & 0.058 & 8.8 & 0.058 \\
        \lastrow A1 & \samel & 5.5 & 0.085 & 5.5 & 0.075 & 5.5 & 0.090 & 17.2 & 0.124 & 28.4 & 0.136 & 21.8 & 0.118 \\
        \lastrow A2 & \samel & 5.5 & 0.065 & 5.5 & 0.070 & 5.5 & 0.070 & 9.4 & 0.086 & 9.2 & 0.080 & 9.4 & 0.084 \\
        \lastrow A3 & \samel & 5.5 & 0.070 & 5.5 & 0.070 & 5.5 & 0.065 & 5.8 & 0.082 & 5.6 & 0.084 & 5.8 & 0.080 \\
        OCNLI & $s_2$ & 10.5 & 0.165 & 10.5 & 0.170 & 10.5 & 0.170 & 14.0 & 0.178 & 14.2 & 0.170 & 14.0 & 0.172 \\
        CAugNLI & $s_1$ & 5.0 & 0.050 & 5.0 & 0.055 & 5.0 & 0.050 & 5.0 & 0.058 & 5.2 & 0.060 & 8.6 & 0.080 \\
        \samel & $s_2$ & 5.0 & 0.050 & 5.0 & 0.050 & 5.0 & 0.050 & 5.0 & 0.050 & 5.0 & 0.050 & 5.0 & 0.050 \\
        SNLI$_\text{debiased}$ & $s_1$ & 6.5 & 0.100 & 6.5 & 0.100 & 6.5 & 0.100 & 6.2 & 0.090 & 6.4 & 0.094 & 6.6 & 0.096 \\
        \samel & $s_2$ & 5.0 & 0.050 & 5.0 & 0.050 & 5.0 & 0.050 & 5.0 & 0.050 & 5.0 & 0.050 & 5.0 & 0.050 \\
        MNLI$_\text{debiased}$ & $s_1$ & 5.5 & 0.065 & 5.5 & 0.065 & 5.0 & 0.065 & 5.8 & 0.072 & 5.8 & 0.078 & 5.8 & 0.074 \\
        \samel & $s_2$ & 5.0 & 0.050 & 5.0 & 0.050 & 5.0 & 0.050 & 5.0 & 0.050 & 5.0 & 0.050 & 5.0 & 0.050 \\
        \midrule
                        &       &     \multicolumn{6}{c}{\textit{k=50}} & \multicolumn{6}{c}{\textit{k=100}}\\
                &       &      \multicolumn{2}{c}{No PCA} & \multicolumn{2}{c}{$|PC|=50$} & \multicolumn{2}{c}{$|PC|=100$} & \multicolumn{2}{c}{No PCA} & \multicolumn{2}{c}{$|PC|=50$} & \multicolumn{2}{c}{$|PC|=100$} \\
        Dataset & Cond.  & L2 & KLD & L2 & KLD & L2 & KLD & L2 & KLD & L2 & KLD & L2 & KLD \\
        \midrule
        SICK & $s_2$ & 6.2 & 0.081 & 7.4 & 0.082 & 7.9 & 0.081 & 13.5 & 0.104 & 13.2 & 0.098 & 12.4 & 0.102 \\
        SNLI & $s_2$ & 11.3 & 0.141 & 9.8 & 0.137 & 9.8 & 0.141 & 12.5 & 0.148 & 10.0 & 0.143 & 10.4 & 0.136 \\
        MNLI-b & $s_2$ & 5.4 & 0.066 & 5.4 & 0.066 & 5.3 & 0.064 & 5.4 & 0.069 & 5.4 & 0.068 & 6.3 & 0.069 \\
        MNLI-u & \samel & 7.4 & 0.073 & 7.4 & 0.076 & 5.9 & 0.078 & 8.5 & 0.087 & 9.3 & 0.087 & 8.6 & 0.086 \\
        XNLI & $s_2$ & 5.5 & 0.074 & 5.6 & 0.074 & 5.6 & 0.073 & 5.7 & 0.079 & 5.7 & 0.080 & 5.8 & 0.079 \\
        FEVER & $s_1$ & 14.6 & 0.154 & 14.7 & 0.152 & 12.7 & 0.149 & 16.7 & 0.159 & 17.2 & 0.157 & 17.8 & 0.160 \\
        ANLI & $s_2$ & 25.2 & 0.132 & 8.9 & 0.063 & 7.0 & 0.058 & 26.9 & 0.162 & 9.2 & 0.071 & 9.0 & 0.072 \\
        \lastrow A1 & \samel & 37.8 & 0.180 & 31.0 & 0.168 & 36.1 & 0.179 & 68.2 & 0.354 & 56.8 & 0.301 & 67.9 & 0.341 \\
        \lastrow A2 & \samel & 17.3 & 0.114 & 19.6 & 0.112 & 15.1 & 0.103 & 37.1 & 0.171 & 31.8 & 0.188 & 34.9 & 0.179 \\
        \lastrow A3 & \samel & 13.4 & 0.108 & 13.8 & 0.106 & 9.8 & 0.094 & 26.8 & 0.147 & 30.5 & 0.162 & 27.8 & 0.152 \\
        OCNLI & $s_2$ & 18.8 & 0.190 & 14.9 & 0.180 & 14.5 & 0.176 & 25.2 & 0.193 & 26.5 & 0.207 & 21.4 & 0.186 \\
        CAugNLI & $s_1$ & 5.0 & 0.051 & 5.0 & 0.050 & 5.0 & 0.051 & 5.9 & 0.059 & 5.9 & 0.056 & 5.1 & 0.056 \\
        \samel & $s_2$ & 13.8 & 0.073 & 6.9 & 0.068 & 12.1 & 0.101 & 19.7 & 0.126 & 19.0 & 0.135 & 13.2 & 0.091 \\
        SNLI$_\text{debiased}$ & $s_1$ & 5.0 & 0.050 & 5.0 & 0.051 & 5.0 & 0.050 & 5.0 & 0.051 & 5.0 & 0.051 & 5.0 & 0.051 \\
        \samel & $s_2$ & 5.9 & 0.080 & 6.3 & 0.088 & 6.1 & 0.085 & 6.0 & 0.085 & 6.9 & 0.090 & 6.2 & 0.086 \\
        MNLI$_\text{debiased}$ & $s_1$ & 5.0 & 0.052 & 5.0 & 0.052 & 5.0 & 0.052 & 5.1 & 0.054 & 5.1 & 0.054 & 5.1 & 0.056 \\
        \samel & $s_2$ & 5.6 & 0.070 & 5.6 & 0.068 & 5.5 & 0.069 & 6.3 & 0.071 & 5.3 & 0.065 & 5.6 & 0.070 \\
        \bottomrule
        \end{tabular}
    \caption{
    PECO scores for various levels of principal component reduction ($|PC|$), various numbers of $k$-means clusters, using L2 and KLD distance. These numbers are collected from a separate reproduction to \autoref{tab:results}, which uses different $k$ for different datasets (reflecting their different sizes), and $|PC|=30$. This table clearly illustrates that e.g., L2 PECO are pretty consistent across different numbers of principal components, for a given number of clusters $k$. We find that calibrating $k$ based on the number of test-set samples for a given dataset is valuable for producing good characterization of the degree of bias. To improve legibility, we kept $\text{PECO}_{KLD}$ in the $[0,1]$ range.
    }
    \vspace{-2ex}
    \label{tab:full_hparam_1}
\end{table*}

\end{document}